\documentclass[a4, 10 pt, conference]{ieeeconf} 
\usepackage{graphicx}
\usepackage{subfig}
\usepackage{amsmath}
\usepackage{multirow}
\usepackage{float}
\usepackage{amssymb}
\usepackage{xcolor}
\usepackage{balance}

\IEEEoverridecommandlockouts                    

\title{\LARGE \bf Effective Target Aware Visual Navigation for UAVs}

\author{Ciro Potena, Daniele Nardi and Alberto Pretto
	\thanks{This work has been supported by the European Commission under the grant number H2020-ICT-644227-FLOURISH. Potena,Nardi and Pretto are with the Department of Computer, Control, and Management Engineering ``Antonio Ruberti``, Sapienza University of Rome, Italy. Email: {\tt \{potena, nardi, pretto\}@diag.uniroma1.it}.}.}

\begin{document}
	
	\maketitle
	\thispagestyle{empty}
	\pagestyle{empty}

\begin{abstract}
In this paper we propose an effective vision-based navigation method that allows a multirotor vehicle to simultaneously reach a desired goal pose in the environment while constantly facing a target object or landmark. Standard techniques such as Position-Based Visual Servoing (PBVS) and Image-Based Visual Servoing (IBVS) in some cases (e.g., while the multirotor is performing fast maneuvers) do not allow to constantly maintain the line of sight with a target of interest. Instead, we compute the optimal trajectory by solving a non-linear optimization problem that minimizes the target re-projection error while meeting the UAV's dynamic constraints. The desired trajectory is then tracked by means of a real-time Non-linear Model Predictive Controller (NMPC): this implicitly allows the multirotor to satisfy both the required constraints. We successfully evaluate the proposed approach in many real and simulated experiments, making an exhaustive comparison with a standard approach. 

\end{abstract}
\section{Introduction}
Vision-based control, or \textit{visual servoing} (VS), of UAVs (Unmanned Aerial Vehicles) is an active research topic with many applications, including search \& rescue, fire monitoring, traffic monitoring and patrolling. More specifically, in these tasks the multirotor is steered to its desired state by using visual feedbacks obtained from one or more cameras. 
This topic has gained even more interest in the last years, making it possible to deal with complex vision-based tasks, such as landing on moving platforms \cite{Lee_2012}, flight through gaps \cite{Falanga2017}, object grasping \cite{Thomas_2014} and target tracking \cite{Pestana_2014}. 

What makes VS a challenging problem for multirotors is the under-actuated dynamics of such vehicles, especially when performing agile and fast maneuvers (i.e., with high velocity and angular accelerations). During such kind of maneuvers, standard visual-based controllers focus solely on reaching the goal state \textit{without} constantly taking into account their camera configuration with respect to the perceived environment.
In other words, during the UAV flight these systems may lose for some time the line of sight with a target of interest, even if such target represents the final goal of the flight. This behavior can prevent the applicability of such controllers in activities when the re-localization of the target of interest is not a trivial task, due to the self-motion of the target (e.g., when tracking a specific person that moves in the crowd), or due to sensor aliasing (e.g., when moving toward a specific object with not unique appearance features).\\
\begin{figure}[t!]
	\centering
 			{\includegraphics[width=\columnwidth]{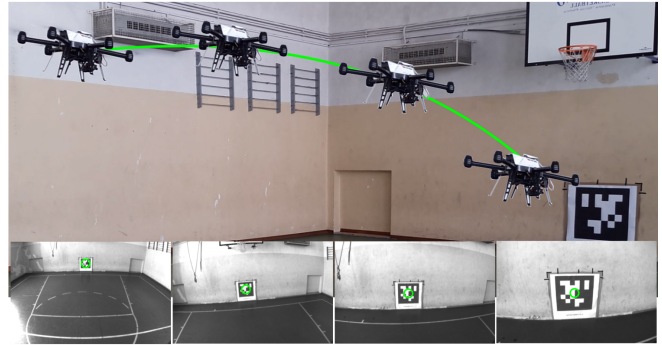}}   	
	\caption{\small An example of target aware visual navigation: the UAV is following an optimal trajectory towards the target  while constantly framing the target with the camera.}
	\label{fig:real_mpcovs}
\end{figure}

In this paper, we propose an effective and robust VS controller that allows an UAV to perform fast maneuvers without losing the line of sight with the target of interest during the entire duration of the flight.\\
VS techniques can be split into two parallel branches: Position-Based Visual Servoing (PBVS) and Image-Based Visual Servoing (IBVS).
In PBVS, the 3D goal pose is directly obtained from a complete 3D reconstruction of the surrounding environment or from the 6D position of one or more landmarks placed in it. In contrast, IBVS formulates the problem in terms of image features locations: the goal pose is defined by means of desired features locations in the final image while the control law aims to minimize the features re-projection error during the flight. Even if IBVS does not require any full 3D estimation, it still needs the depth of the target.
It has been shown that both strategies have their own weakness. In IBVS it is particularly challenging to model the relation between the vehicle dynamics and the feature projection error, especially for under-actuated systems. Furthermore, an inaccurate estimation of the object depth leads to instabilities.
In PBVS, since the control law is directly designed in the state-space domain, there is a close dependence on the accuracy of the 3D environment reconstruction or on the target pose estimation. In practice this estimation may be very noisy, leading PBVS to be very sensitive to initial conditions, camera calibration parameters and image noise corruption.
 
Differently from the previous work, we propose a procedure that decouples the planning and the control problems. 
The planning task is addressed by employing a hybrid approach. Firstly, as in PBVS, we get the goal pose as the position and the relative orientation of the vehicle in the environment that allows to have the desired view of the target object.

Then, similarly to IBVS, we model the trajectory as a non-linear constrained optimization problem with a cost function that penalizes the target's location error, in order to constantly keep the target in the camera field of view.

Once a global optimal trajectory\footnote{With an abuse of notation, as in other related work, we call here and in the rest of the paper "optimal trajectory" the desired trajectory the multirotor tracks during the flight. Actually, due to the non-linear nature of the cost function, the optimization does not always guarantee the convergence to the optimal, global minimum.} has been found, we employ an NMPC framework as controller and local planner. Making use of an efficient open-source solver, our control framework is capable to solve an NMPC problem in few milliseconds, allowing us to use at each time step just the initial tuple of control inputs\footnote{An NMPC provides a sequence of control inputs for a finite temporal horizon.} while simultaneously re-solving the whole non-linear control problem.\\

We compare our method against a common PBVS approach in both simulated and real environments, getting in all experiments cutting edge results. Additionally, we make a preliminary assessment with respect to a state-of-the-art Optimal Visual Servoing (OVS) technique, suggesting that our approach can achieve comparable results. 

\subsection{Related Work}

Several IBVS \cite{Pestana_2014}\cite{hamel_2002}\cite{Guenard_2007} and PBVS \cite{Altug_2002}\cite{Mejías_2006} approaches has been applied to control aerial vehicles in the last decades. In those standard solutions, the controller uses the visual information as the main source for the target pose computation, \textit{without} taking into account where the target is re-projected into the image plane along the trajectory. A possible solution that usually mitigates such weakness is the kinematic limitation of the multirotor in terms of roll and pitch angles, but this penalizes the vehicle maneuverability. To this end, Ozawa \textit{et al.} \cite{Ozawa_2011} present an approach that takes advantage of the rotational-dynamic of the vehicle, where a virtual spring penalizes large rotation with respect to a gravity aligned frame.
Some recent approaches map the target features' dynamic into a "virtual image-plane" used to compensate the current roll and pitch angles, in order to keep them close to zero \cite{Lee_2012}\cite{Thomas_2014}\cite{Jabbari_2014}. Being the re-projection error obtained from the rotation-compensated frame, it is still possible that the target, due to significant rotations, completely leaves the camera field of view. 

In \cite{Bourquardez_2009}\cite{Mebarki_2015} the authors present two approaches based on a spherical camera geometry, allowing to design the control law as a function of the position while neglecting the angular velocity. Being solely position-based, these kind of methods suffer from the above discussed problems, since the system is still vulnerable to large rotations.

Some recent approaches are based on hybrid techniques, where image features and 3D data are fused together to develop a more stable controller than IBVS or PBVS alone.  An example has been presented in \cite{Hafez_2008}, where the outputs from IBVS and PBVS methods are fused to form a stable Hybrid controller.  Sheckells \textit{et al.} \cite{Sheckells_2016} presented an approach where the desired trajectory is obtained by minimizing a cost function over the re-projection error. The proposed optimization procedure leads to computational time constraints that do not allow to constantly re-optimize the whole path while following it, penalizing the vehicle to obtain even more better tracking performance.
Our work builds on a part of the problem formulation given in \cite{Sheckells_2016}, but our solution presents significant differences: (i) the whole problem is decoupled and split into two optimization problems; (ii) The formulation of the target re-projection error assumes a slightly different form, by enabling to scale the different error components; (iii) the NMPC explicitly takes into account two dynamic effects and the low-level controller that runs on the UAV.\\ 
The coupling between perception and planning has also been addressed in \cite{Falanga2017}, where an UAV has to simultaneously localize itself with respect to a gap and pass trough it. They plan a ballistic trajectory capable to satisfy both dynamic and perception constraints by maximizing the distance of the vehicle with respect to the edges of the gap.\\

In all the above mentioned work, except for \cite{Sheckells_2016}, there is no guarantee that the target is constantly kept in the camera field of view because they don't directly take into account the vehicle dynamics. 

\subsection{Contributions}

Our method differs from previous works under two main aspects: (i) Unlike standard VS approaches, we guarantee to constantly maintain the target as close as possible to the center of the camera field of view during the whole maneuver; (ii) Making use of a global and a local planner, we allow the multirotor to constantly stay on the optimal path.

\section{UAV Dynamic Model} \label{sec:dynamic_model}
In this section, we describe the vehicle dynamic model that we exploit as a constraint in the optimal trajectory computation.\\
We express a generic position vector as $x^Z_Y$ denoting the position of the reference frame $Y$ expressed with respect to the reference frame $Z$. Furthermore, we express a rotation matrix from the reference system $Y$ to the reference system $Z$ as $R^Z_Y$.
For the trajectory planning and the control of the multirotor vehicle, we make use of three main coordinate reference systems: 
(i) the camera frame with $C$; (ii) the world fixed inertial frame with $I$; (iii) the body fixed frame with $B$, that is the frame attached to the Center of Gravity (CoG) of the UAV. The UAV configuration at each time step is formulated by the position $p^I_B$ and the linear velocities $v^I_B$ of the vehicle CoG, both expressed in the inertial frame, and the vehicle orientation $q^I_B$. More specifically, the whole state of the vehicle is then expressed as $x=\{p^I_B, q^I_B, v^I_B\}$. At each time step, we also define the tuple of control inputs as $u=\{ \phi_{cmd}, \theta_{cmd}, \dot\psi_{cmd}, T_{cmd}\}$, where the single terms stands for, respectively, the roll, pitch and yaw rate desired commands and the commanded thrust.

We employ a widely used dynamic model for multirotors, where the main forces that act on the vehicle are generated from the propellers. More specifically, each propeller generates a thrust force $F_T$ proportional to the square of the motor rotation speed. Moreover, we take into account also two other important effects that became relevant in case of dynamic maneuvers, namely \textit{blade flapping} and induced \textit{drag}. Both of them introduce additional forces in the x-y rotor plane \cite{Mahony_2012}.
We model them into a single lumped \textit{drag} coefficient $K_D$, as shown in \cite{Omari_2016, BUrri_2012}, leading to the aerodynamic force $F_{aero,i}$:

\begin{equation}
  F_{aero,i} = F_{T,i}K_{drag}{R^I_B}^Tv^I_B
\end{equation}

where $i$ stands for the propeller index, $K_{drag} = diag\{K_D, K_D, 0\}$, $F_{T,i}$ is the $z$ component of the $i-th$ thrust force and $v^I_B$ is the vehicle's linear velocity (in the next equations, where there is no confusion, we will omit both the superscripts and subscripts $I$ and $B$).
The final dynamic model of the vehicle can be expressed as follows:

$$
\dot{p} = v, \eqno{(2.a)}
$$
$$
\dot{v} = \frac{1}{m} \Big( R \sum_{n=0}^{n_p}( F_{T,i}- F_{aero,i} ) + F_{ext} \Big) + g , \eqno{(2.b)}
$$
$$
\dot{\phi} = \frac{1}{\tau_{\phi}}(k_{\phi}\phi_{cmd} - \phi) \eqno{(2.c)}
$$
$$
\dot{\theta} = \frac{1}{\tau_{\theta}}(k_{\theta}\theta_{cmd} - \theta) \eqno{(2.d)}
$$
$$
\dot{\psi} = \dot\psi_{cmd} \eqno{(2.e)}
$$

where $m$ is the mass of the vehicle, $F_{ext}$ are the external forces that act on the multirotor.
In our system, we make use of a low-level controller that maps the high-level attitude control inputs in propellers' velocity, as the one provided with the  Asctec NEO hexacopter used in the experiments.
To achieve better tracking performance, we model this inner control loop as first order dynamic systems, where the model parameters $\tau_i$ and $k_i$ are obtained by a system identification procedure \cite{Kamel_2016}.

\section{Optimal Visual Servoing (OVS)}\label{sec::ovs}
	
In this section we describe how we take into account dynamics and perception constraints in planning a trajectory and controlling the multirotor. 
The first step is the goal pose computation, namely the position and orientation of the vehicle that allows to get the desired view of the target. We then split the Optimal Visual Servoing (OVS) problem into two consecutive stages. 
First, we compute an optimal global trajectory by solving a non-linear optimization problem. In order to take into account the dynamic and perception constraints, the output trajectory is minimized over the multirotor dynamics and the target re-projection error in the image plane.
To track the desired trajectory we then employ a Receding Horizon NMPC controller, where a smaller non-linear optimization problem is solved every time step and only the first control input is actually sent to the multirotor.

\subsection{Goal Pose Computation}

Before computing the optimal trajectory, the multirotor has to retrieve the goal pose it aims to reach. Such a pose depends on the task (e.g., inspection or patrolling) and it usually requires the vehicle to frame a target (e.g., landmark or object) from a specific distance and with a specific point of view. Retrieving a relative 3D transformation from the camera is a well-known problem and has been widely studied in the last decades. A widely used technique is based on the solution of a Perspective-n-Point ($PnP$) problem \cite{Petersen_2008}: such technique requires a prior knowledge about the target object geometry and scale. 

Since the choice of the goal pose computation algorithm goes behind the purpose of this work, we assume for the sake of simplicity to have a real-time ''black-box'' detection framework that outputs: (i) the $(u,v)$ pixel coordinates of the target $T$ in the camera image plane; (ii) the 3D position of the target in the camera frame $p_T^C$; (iii) the orientation $q_C^T$ of the target object with respect to the camera frame $C$ in terms of \textit{yaw} angle. 
The goal pose in world $I$ reference system can be then obtained as follows:

$$
{p_{goal}}_B^I =  p_B^I + q_B^I(  q_C^B( d_T^C - p_T^C  ) + p_C^B) \eqno{(3.a)}
$$
$$
{q_{goal}}_B^I = q_B^I q_C^B q_T^C   \eqno{(3.b)}
$$

Where $d_T^C$ is the desired position of the target expressed in the camera frame, while $p_C^B$ and $q_C^B$ are the extrinsic calibration parameters between the camera frame and the body frame.

\subsection{Optimal Trajectory Computation}

Once we have the goal pose, we need to generate a discrete trajectory composed by $N$ tuples of the vehicle state vector $\{ x_0, ..., x_N \}$ and control inputs $\{ u_0, ..., u_N \}$ that minimize the functional cost $J$ subject to the vehicle dynamics equations $f(x_k, u_k)$ described in Sec.~\ref{sec:dynamic_model}. The time step of such dynamic equations is given by $\frac{t_f-t_0}{N}$, where $t_f$ and $t_0$ are respectively the final time and the initial time, while $N$ is the number of steps. Additionally, the custom choice of the time variable allows us to define also the nominal speed $s_{nom}$ (i.e. $\frac{\Delta_p}{t_f}$), namely the speed the vehicle is expected to flight.
Similarly to \cite{Sheckells_2016}, we define the cost function as:
$$
J( x_{0:N},u_{0:N-1} ) = J_{N}(x_{N}) + \sum\limits_{k=0}^{N-1}J_{k}(x_k,u_k)    \eqno{(4.a)}
$$

where $J_{N}$ is the final cost and $J_{k}$ is the cost along the trajectory. At this point, we split $J_{k}$ into two main terms. The first one represents the cost over the desired final state and the control effort, and it can be expressed as follow:

$$
J'_{K}(x_k,u_k) = \frac{1}{2}(x_k-x_N)^TQ(x_k-x_N) + \frac{1}{2}u^TRu \eqno{(4.b)}
$$

where $Q\geq0$ and $R\geq0$ are the matrices that weight the control objectives. In addition, in the second term of $J_{k}$ we introduce a cost that aims to penalize the re-projection error of the target into the camera field of view. The entire cost in the discrete time step $k$ can be then formulated as: 

$$
J_{K}(x_k,u_k) = J'_{K}(x_k,u_k) + \frac{1}{2}e_i(x_k)^THe_i(x_k)\eqno{(4.c)}
$$
$$
e_i(x_k) = \mathcal{P}(x_k, P_i, \pi) - p_i \eqno{(4.d)}
$$

where $H\geq0$ is the penalization term over the target re-projection error and $\mathcal{P}$ is a general camera projection function. Starting from the 3D position of the object in the camera frame $P_i$, the re-projection error is obtained by the knowledge of the intrinsic calibration parameters of the camera, denoted in Eq. 4.d as $\pi$, the extrinsic parameters between the camera reference system $C$ sensor and the body frame $B$, and the desired position of the target object in the image plane $p_i$. Differently from \cite{Sheckells_2016}, we make use of a weighting matrix H in place of a scalar weighting factor, allowing us to scale the different components of the re-projection error. Ideally, we want to have an $H$ that penalizes mostly the error along the smaller dimension of the input image. We set $H$ as follows:

$$
H = \left( \begin{matrix} h_x \hspace{.8cm} 0 \\ 0 \hspace{.8cm} h_y \end{matrix} \right)  \eqno{(4.e)}
$$

Let $h_{i=x,y}$ be the scale factor related to the smaller dimension $(d_i<d_j)$, we set it as follows:

$$
h_i = h_j\times\sigma,\hspace{.3cm} \sigma = \frac{d_j}{d_i} \eqno{(4.f)}
$$

This enables the UAV to cope with different camera sensor setups. 
Since we introduce in the cost function $J$ the re-projection error term of the target with respect to the camera image plane, the optimal solution will implicitly allows the vehicle to constantly face the target, maintaining it as close as possible to the center of the image plane.

\subsection{Optimal Control Solver}

Once the optimal trajectory has been obtained, the multirotor must closely follow it. To this end we employ an NMPC that repeatedly solves the following optimal control problem:

$$
\min\limits_{u,x} \sum\limits_{k=0}^{K-1} \bigl( \|x_k-x_f\|^2_Q + \|u_k-u_f\|^2_R \bigr) + \|x_K-x_f\|^2_P  \eqno{(5)}
$$

subject to: $ x_{k+1} = f(x_k,u_k)+f_{ext}(d_k)$ \

\qquad \qquad \qquad $ d_{k+1} = d_k $\

\qquad \qquad \qquad $ U_{min} \leq u_k \leq U_{max} $\

\qquad \qquad \qquad $ x_0 = x_{init} $	\

where $Q\geq0$ is the weight factor over the state, $R\geq0$ is the weight factor over the control inputs and $P$ is the weight factor over the final state. The controller is implemented in a receding horizon fashion, meaning that the aforementioned optimization problem is solved every time step over the fixed time interval $[i, i + K]$. 
Once the optimization problem has been solved, the optimization procedure is repeated for the time interval  $[i + 1, i + K + 1]$ starting from the state reached in $i+1$ and by using the previous solution as initial guess.
By solving this optimization procedure in real-time, the proposed framework simultaneously provides a feed-forward trajectory toward the desired state and a discrete set of control inputs which will be used by the low-level on-board controller. This means that, in practice, at the end of each optimization procedure only the first control input tuple is actually sent to the multirotor controller, then the optimization procedure is repeated.

\section{Simulation Experiments}\label{sec:experiments}

\begin{figure*}[ht]
		\centering
		\begin{tabular}{cc}
			{\includegraphics[width=0.45\textwidth]{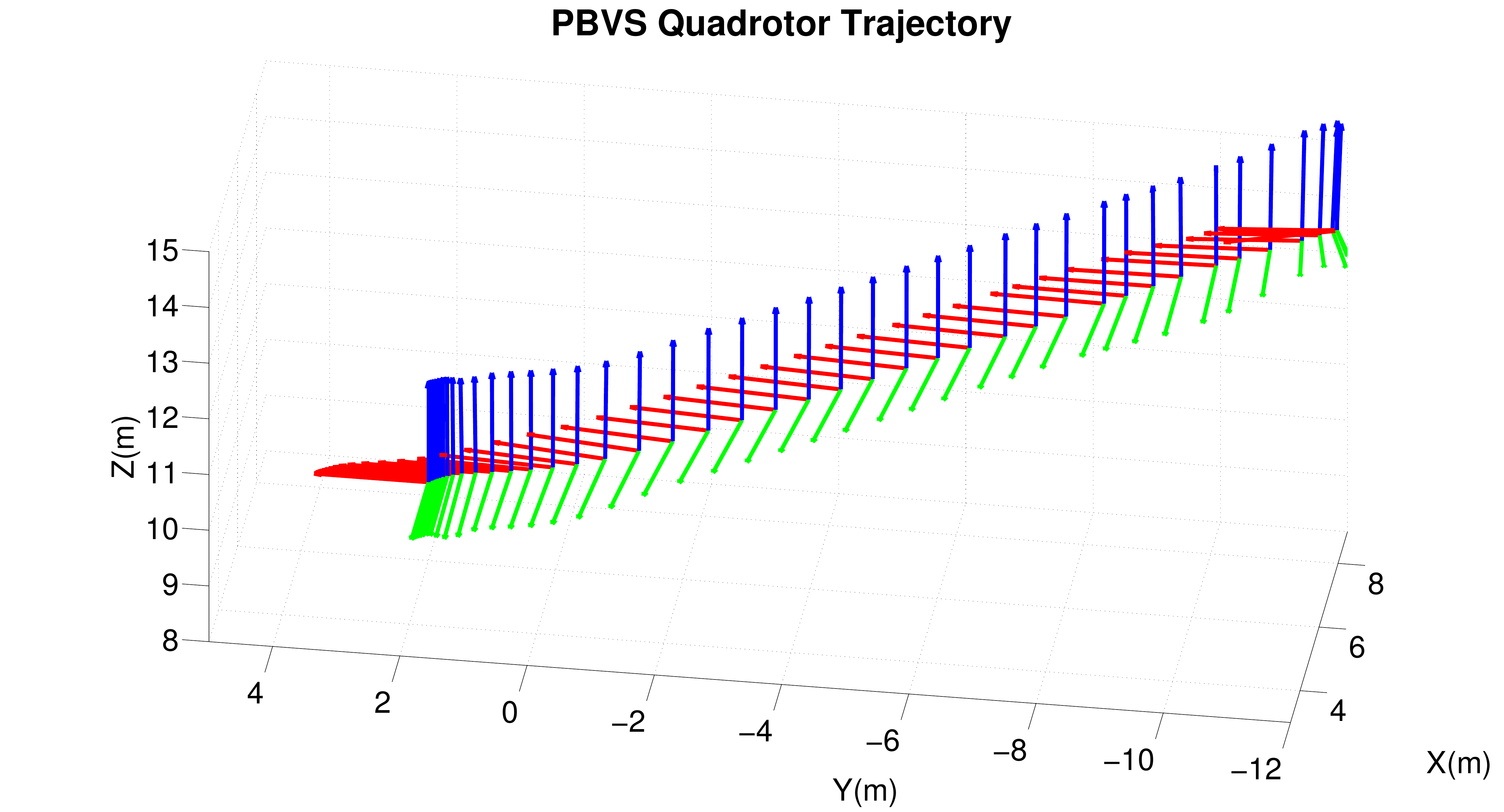}} 
			{\includegraphics[width=0.45\textwidth]{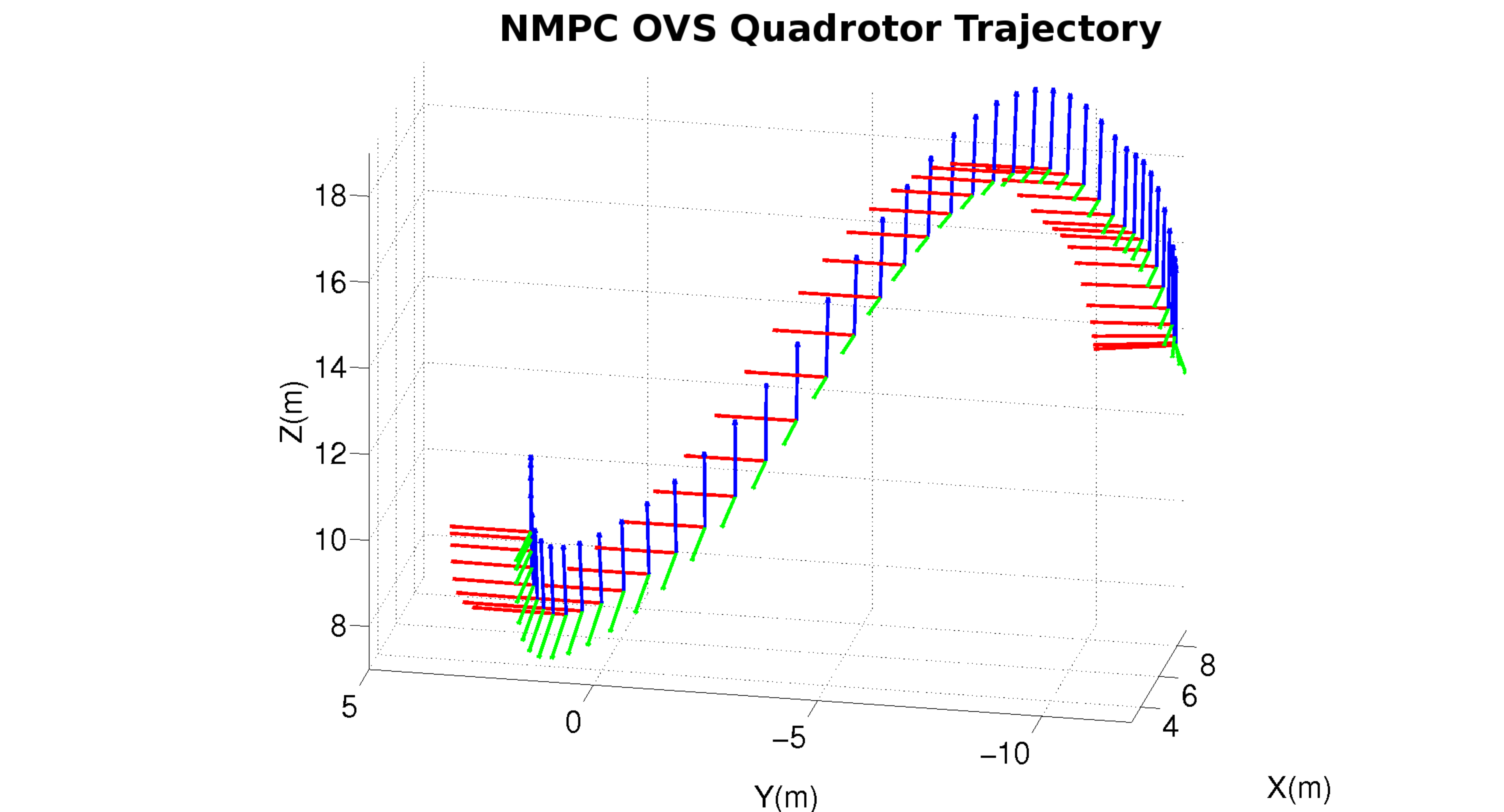}}    
		\end{tabular}
		
		\caption{\small Example of trajectories obtained using PBVS(a) and NMPC OVS (b) with $s_{nom} = 3.10\frac{m}{s}$: the latter constantly takes into account the target pose during the flight.\vspace{.5cm} }
		\label{fig:trajectories}
\end{figure*}

\begin{table*}[ht]
		
     		\centering
     		\caption[caption of PES]{ Comparison of simulated image error trajectory statistics for each method across different nominal speeds\protect\footnotemark.}
     		\label{tab:sim_err_stats}
     		\begin{tabular}{ llllllll }
     			\hline
     			\multicolumn{2}{c}{} & \multicolumn{3}{c}{Avg. Pixel Error} & \multicolumn{3}{c}{Max. pixel error}\\
     			\cline{3-8	}
     			$t_f$ & $S_{nom}$  & NMPC OVS 	      	      & PBVS  &Sheckells \textit{et al.}\cite{Sheckells_2016}\footnotemark[\value{footnote}]   & NMPC OVS & PBVS   &Sheckells \textit{et al.}\cite{Sheckells_2016}\footnotemark[\value{footnote}]\\\hline
     			10.2  & 2.01       & \textbf{32.5}    	      & 89.2  & $\sim$63.8 		 & \textbf{55.05}   & 145.36  & $\sim$127.6		     \\ 
     			7.5   & 2.73       & \textbf{45.03}   	      & 109.4  & $\sim$85.3		 & \textbf{76.91}   & 199.21 & $\sim$195.1		     \\
     			6.6   & 3.10       & \textbf{55.2}    	      & 125.2  & $\sim$90.7		 & \textbf{98.76}   & 223.67 & $\sim$207.9		     \\
     			5.1   & 4.02       & \textbf{62.5}    	      & 146.9 & not available		 & \textbf{115.64}  & 323.78 &  not available		     \\\hline\vspace{.1cm}

     		\end{tabular}
\end{table*}

\begin{table*}[ht]
	  
	  \centering
	  \caption{ Comparison in terms of control effort between a standard PBVS approach and the proposed one. }
	  \label{tab:control_efforts}
	  \begin{tabular}{ llllllllll }
		  \hline
		  \multicolumn{2}{c}{} & \multicolumn{2}{c}{ RMS Thrust (g)} & \multicolumn{2}{c}{ RMS Roll Ref. (deg)} & \multicolumn{2}{c}{ RMS Pitch Ref. (deg)} & \multicolumn{2}{c}{ RMS Yaw Rate (rad/s)}\\
		  \cline{3-10}
		  $t_f$ & $S_{nom}$  & NMPC OVS & PBVS  & NMPC OVS & PBVS   & NMPC OVS & PBVS  & NMPC OVS & PBVS  \\\hline
		  10.2  & 2.01       & 10.8    & 10.56 & 0.15    & 0.11  & 0.08    & 0.075  & 0.34    & 0.33 \\ 
		  7.5   & 2.73       & 10.95   & 10.79 & 0.31    & 0.25  & 0.33    & 0.29   & 0.43    & 0.46 \\
		  6.6   & 3.10       & 11.21   & 10.98 & 0.5     & 0.43  & 1.47    & 0.38   & 0.49    & 0.48 \\
		  5.1   & 4.02       & 11.54   & 11.13 & 0.9     & 0.75  & 1.82    & 0.69   & 0.61    & 0.60 \\\hline\vspace{.15cm}

	  \end{tabular}
  \end{table*}
  
We tested the proposed framework firstly in a simulated environment by using the RotorS simulator \cite{Furrer_2016} and a simplified multirotor model with a front-facing camera. The mapping between the high-level control input and the propellers velocities is done by a low-level PD controller that aims to resemble the low-level controller that runs on the real multirotor.
From the higher controller level point of view, we implemented a receding horizon NMPC \cite{neunert_2016}, where the optimization problem is solved by means of the efficient ACADO solver \cite{Houska_2011}.
To demonstrate the effectiveness of the proposed method, we fix the number of segments $N$ and then flew the virtual vehicle to the desired goal pose by using the approach described in section \ref{sec::ovs} and a standard PBVS technique. 
The latter adopts a linear interpolation technique between the starting and the goal poses obtaining a vector of $N$ intermediate poses. Such interpolated poses are then sent to the same NMPC that computes the trajectory to track them. 
Once the final time $t_f$ has been fixed, by tuning $N$ it is possible to act on the flight behavior: 
increasing the number of segments will involve smoother trajectories and control inputs, since the delta-pose between two adjacent desired states segments is smaller. On the other hand, increasing $N$ also brings to higher computational cost when performing the optimal trajectory computation. We used $N = 55$ as trade-off between smoothness and computational velocity.

The goal pose is computed for each run employing an April Marker \cite{Olson_2011} attached on a virtual building. 
Since we aim to test our approach with different levels of aggressive maneuvers, we act on the $S_{nom}$ parameter (i.e. changing the final time $t_f$).
In all the experiments we set the initial state to $x=\{8, -12, 14, 0, 0, 1.918, 0, 0, 0\}$. Since the target is always kept in the same location inside the virtual environment, the computed goal state is $x=\{6+w_x, 2+w_y, 9.4+w_z, 0, 0, 1.57+w_{yaw}, 0, 0, 0\}$, where $w\in\mathbb{R}^4$ is a small white noise random component due to the target detection errors.
The relative transformation between the initial and the final pose forces the multirotor to retrieve an optimal trajectory along the 4 principal motion directions of the vehicle.   

\subsection{Results}

Quantitative image error trajectories for the OVS and PBVS methods for various values of $t_f$ are reported in in Table \ref{tab:sim_err_stats}. The reported results are obtained averaging the performance of PBVS and OVS given the same goal pose and starting from the same initial state for multiple trials. As a preliminary assessment, we also reported some results from the experiments in Sheckells \textit{et al.} \cite{Sheckells_2016}, showing that our method can provide results comparable with this state-of-the-art approach. It is important to highlight that, given this data, a direct comparison with \cite{Sheckells_2016} is not possible, since the pixel error statistics are strictly correlated with the simulation setup which has not been released by the authors.

\begin{figure}[ht]
	\centering
		{\includegraphics[width=0.49\linewidth]{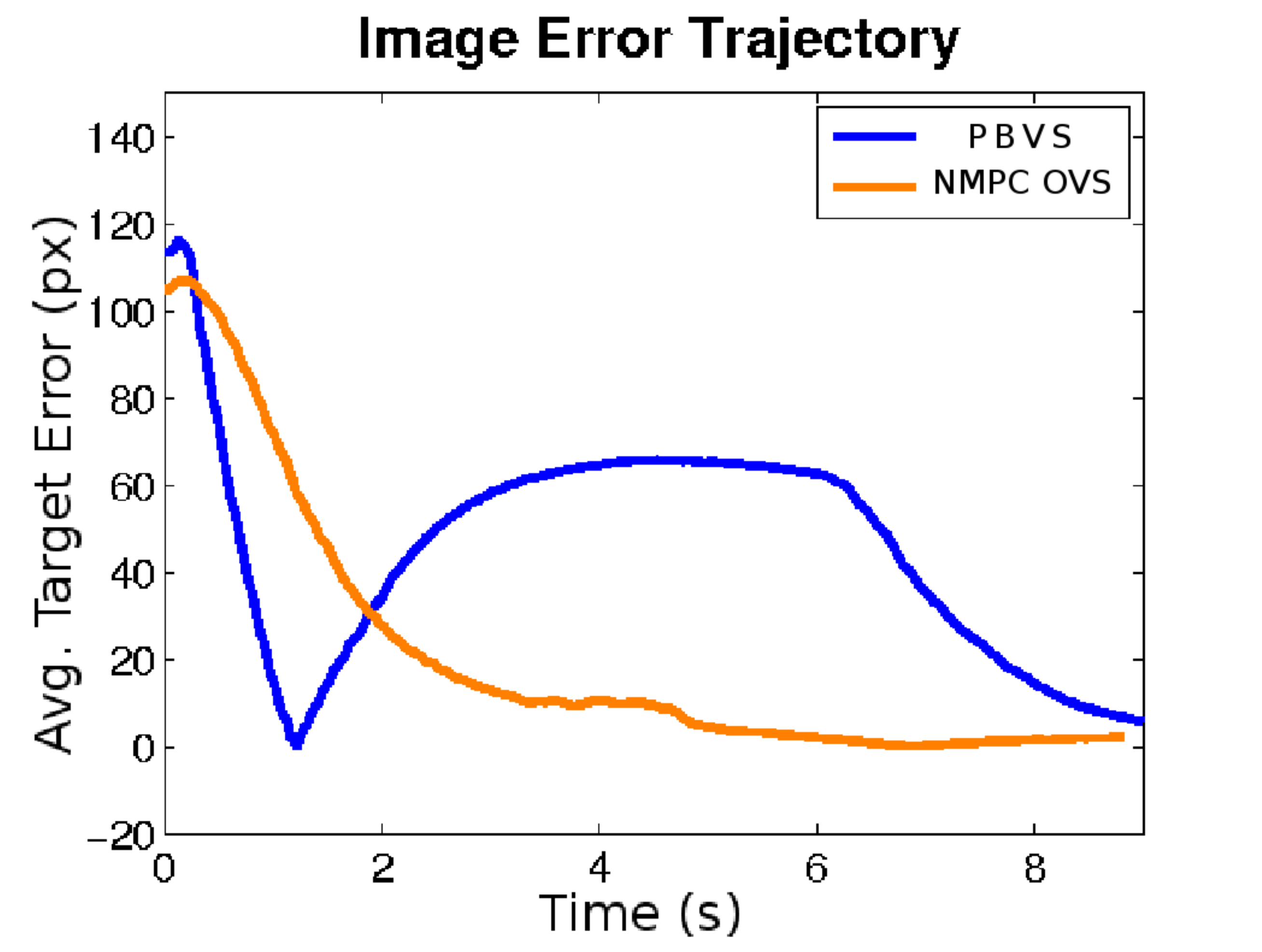}}         
		{\includegraphics[width=0.49\linewidth]{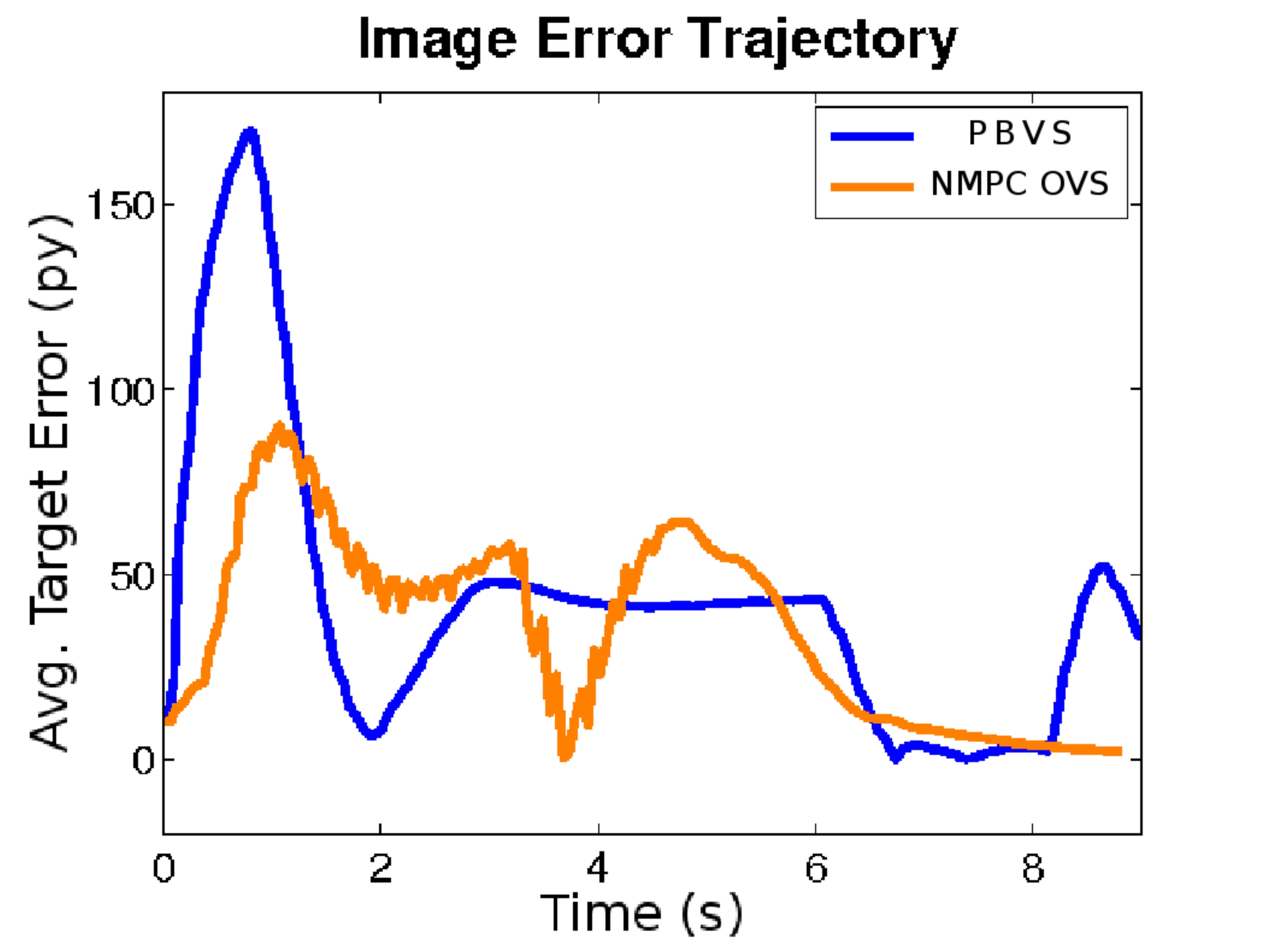}} 
	\caption{\small Comparison of simulated PBVS and NMPC OVS pixel error trajectories for $s_{nom} = 3.10\frac{m}{s}$. Respectively error on the x-axis of the image plane in the left image, while in the right one the pixel error on the y-axis.}
	\label{fig:pbvs_vs_mpcovs}
\end{figure}

Remarkably, the target error trajectory along both image axes is almost always lower than both the other approaches. In spite of this, from a qualitative point of view (see Fig. \ref{fig:pbvs_vs_mpcovs}), the PBVS trajectory seems to behave better in term of pixel errors at same points. The explanation for such behavior comes from the different shape of the two trajectories. In our case, the vehicle is steered to avoid the target to leave the center of the camera field of view, preferring a constant and possibly small error. In the PBVS case, the trajectory is straightforward, involving bigger errors in the acceleration and deceleration phases, worst average and maximum errors, but sporadically smaller error compared with the OVS approach.

\footnotetext{We emphasize that the statistics from \cite{Sheckells_2016} have been obtained with a different simulation setup, so they represent an indicative performance measure.}

\begin{figure}[ht]
	\centering
 			{\includegraphics[width=0.49\linewidth]{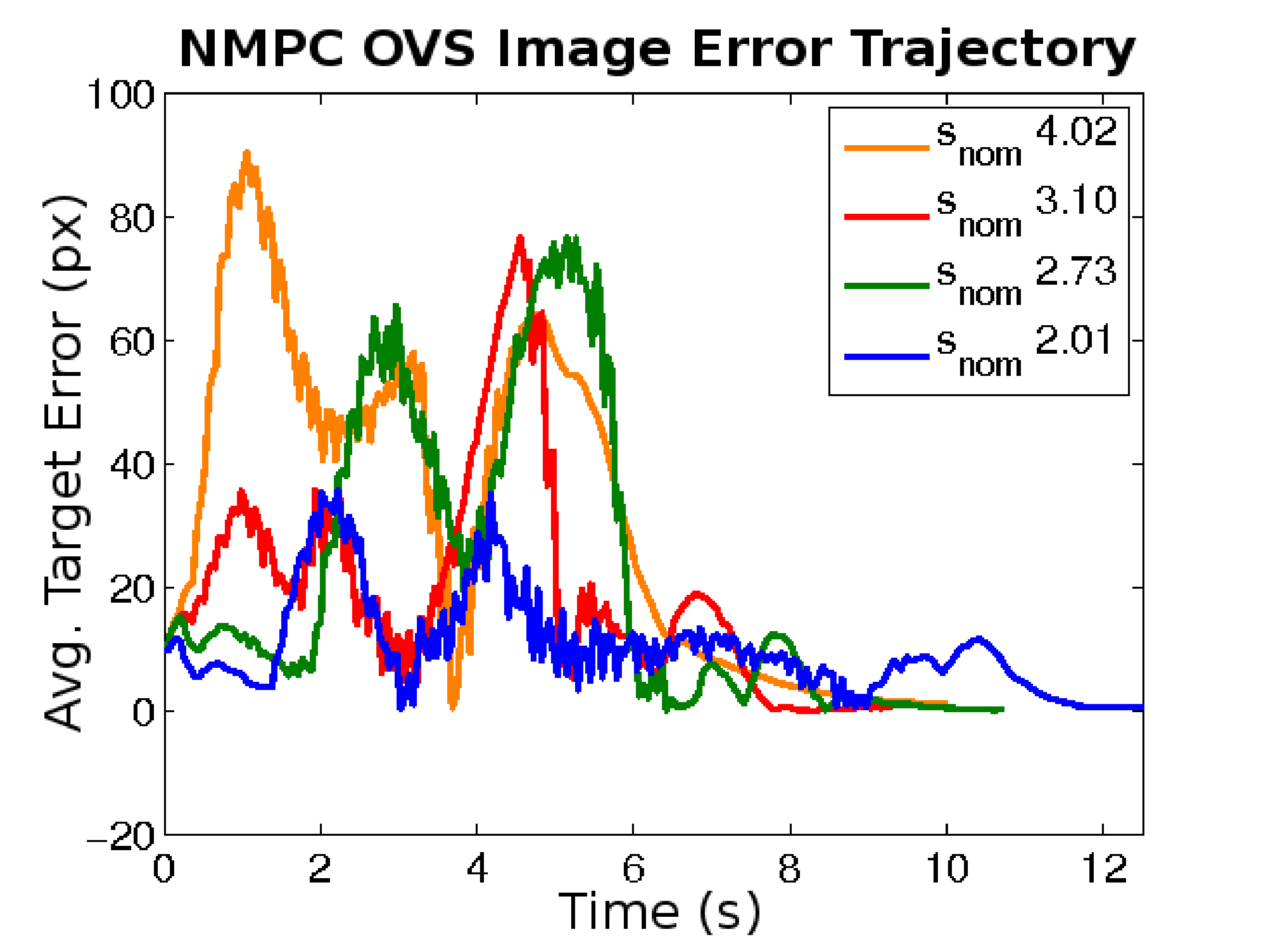}}         
 			{\includegraphics[width=0.49\linewidth]{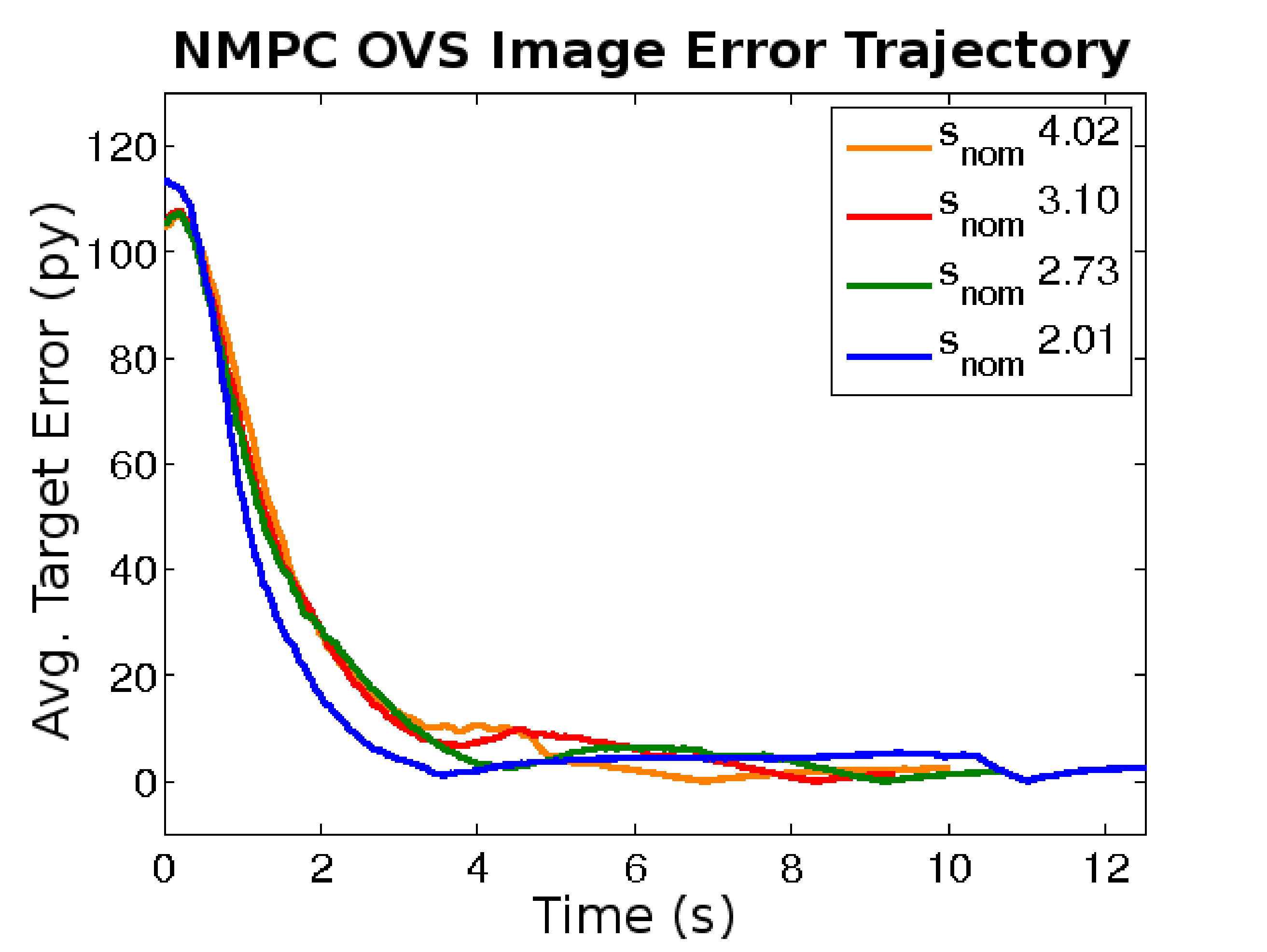}} 
	\caption{\small Comparison of simulated NMPC OVS pixel error trajectories for different values of $s_{nom}$. Respectively error on the x-axis of the image plane in the left image, while in the right one the pixel error on the y-axis.}
	\label{fig:mpcovs}
\end{figure}

From the control inputs point of view, the reduced pixel error comes with an energy effort trade-off: as reported in Table \ref{tab:control_efforts}, the RMS thrust of each OVS trajectory is larger with respect to the corresponding PBVS trajectory. Similar conclusion can be drawn from the attitude points of view since maintaining the target in the center of the camera involves larger angles and yaw rate commands. The choice of the correct behavior depends on the task requirements and, by acting on the optimization parameters $Q_k$, $R_k$ and $H_k$, it is possible to obtain the desired trade-off between control effort and image error. 
Two samples of the trajectories generated by the approaches are depicted in Fig. \ref{fig:trajectories}. Often the bigger pixel error terms in a VS scenario occur in the initial and in the final phase, due to the attitude components required to accelerate and decelerate the vehicle. As qualitatively reported in Fig. \ref{fig:trajectories}, the OVS trajectory takes into account these two error sources by a small ascending phase at the same time as the forward pitch command.
Similarly the trajectory dips softly at the end of the flight so that the target remains in the center of the image plane when the multirotor has to pitch backward in order to decelerate. From the PBVS point of view, the trajectory is more or less a straight line. The vehicle starts suddenly to pitch and to decrease its altitude, involving a bigger pixel error.

\section{Real Experiments}
\begin{figure}[t!]
	\centering
 			{\includegraphics[width=\columnwidth]{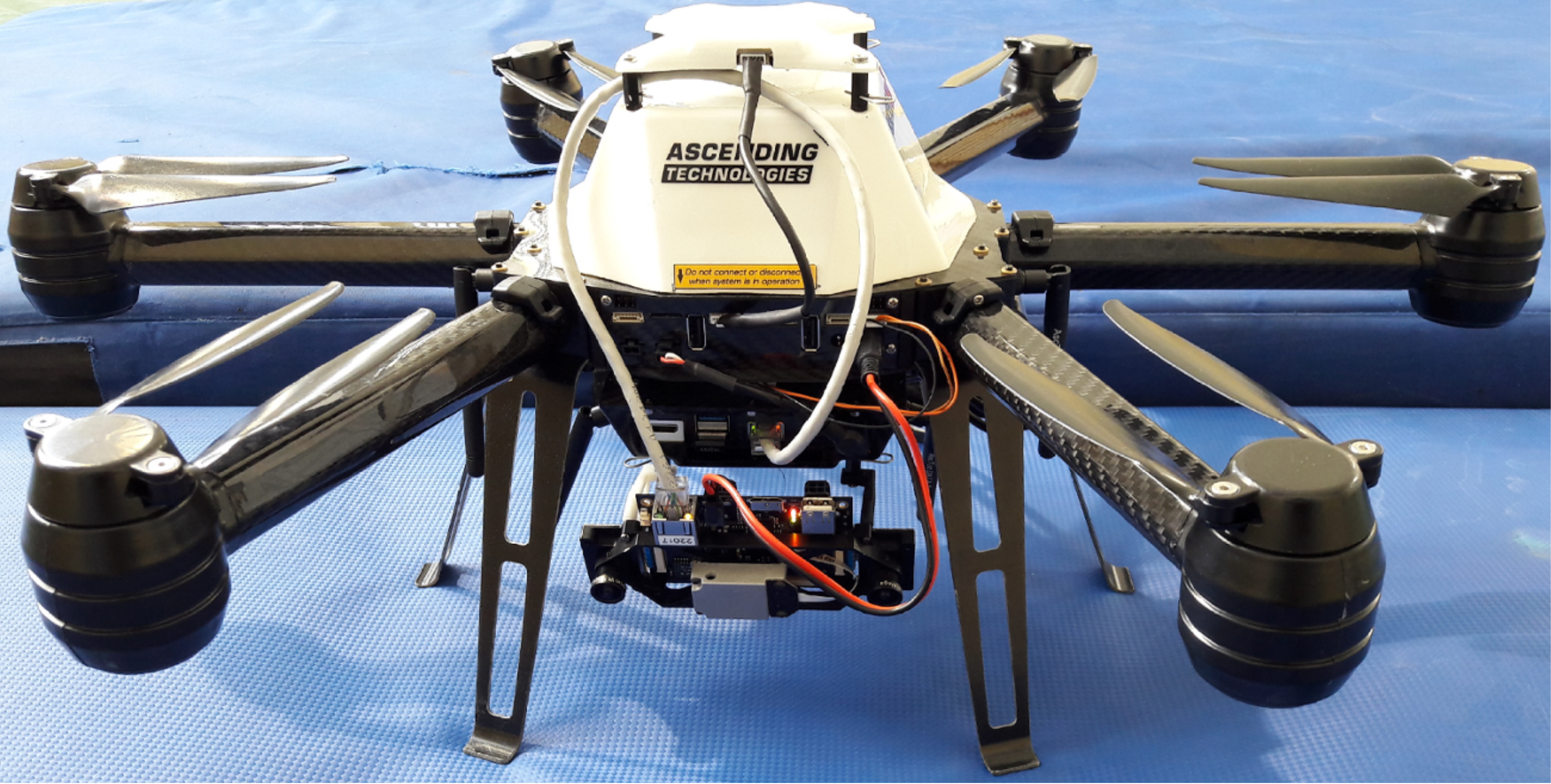}}   	
	\caption{\small The Asctec NEO hexacopter used for the real experiments.}
	\label{fig:neo}
\end{figure}    
We tested the proposed framework on an Asctec NEO hexacopter Fig.~\ref{fig:neo} equipped with an Intel NUC i7, where we implemented our algorithm in ROS (Robotic Operating system), running on Ubuntu 14.04. The overall weight of the vehicle is 2.8 Kg. 
For the state estimation we make use of a forward-looking VI-Sensor \cite{nikolic_2014} and the ROVIO (Robust Visual Inertial Odometry) framework \cite{bloesch_2015}. The ROVIO output is then fused together with the vehicle Inertial Measurement Unit (IMU) by using an Extended Kalman Filter (EKF) as described in Lynen \textit{et al.} \cite{lynen_2013}. The control inputs obtained at each time step by our approach are then sent to the low-level on-board controller by using the UART connection. 

We executed several experiments acting on the $s_{nom}$ parameter. In each run the multicopter starts from the same initial state before starting to look for the fixed target. The distance between the vehicle and the target is approximatively 8 m, and the environment is an indoor closed building. 

\begin{table}[h]
	        
     		\centering
     		\caption{ Comparison of image error for each method across different nominal speeds. }
     		\label{tab:real_err_stats}
     		\begin{tabular}{ llll}
     			\hline
     			\multicolumn{2}{c}{} & \multicolumn{2}{c}{Avg. Pixel Error}\\
     			\cline{3-4}
     			$t_f$ & $S_{nom}$  & NMPC OVS 	      	      & PBVS 	 \\\hline
     			6   & 1.79       & \textbf{62.89}    	      & 77.2  	 \\ 
     			5   & 2.11       & \textbf{88.03}   	      & 112.4  	 \\
     			4   & 2.73       & \textbf{104.17}    	      & 146.7  	 \\
     			3   & 3.21       & \textbf{123.8}    	      & 179.2 	 \\\hline\vspace{.1cm}

     		\end{tabular}
\end{table}
 
\subsection{Results}

Qualitative results for trajectories with different values of $s_{nom}$ are reported in Fig. \ref{fig:real_mpcovs}, while table \ref{tab:real_err_stats} reports the average error statistics between OVS and PBVS. Apart from the same error behavior that also appears in the simulation experiments, it is possible to note how the Mean Squared Error (MSE) pixel error for the OVS approach is lower than the PBVS approach. The difference in terms of pixel MSE is bigger in the initial and final phases where the PBVS, in order to accelerate and slow down, is subject to the greatest kinematic movements in terms of roll and pitch angles.
Is also useful to highlights how both OVS and PBVS use a noisy target detection approach. In practice, the multirotor is not able to obtain an accurate 3D position of the target object in the environment.
This accounts for a constant non-zero pixel MSE, even when the vehicle reach the goal state.

\begin{figure}[h]
	\centering
 			{\includegraphics[width=0.49\linewidth]{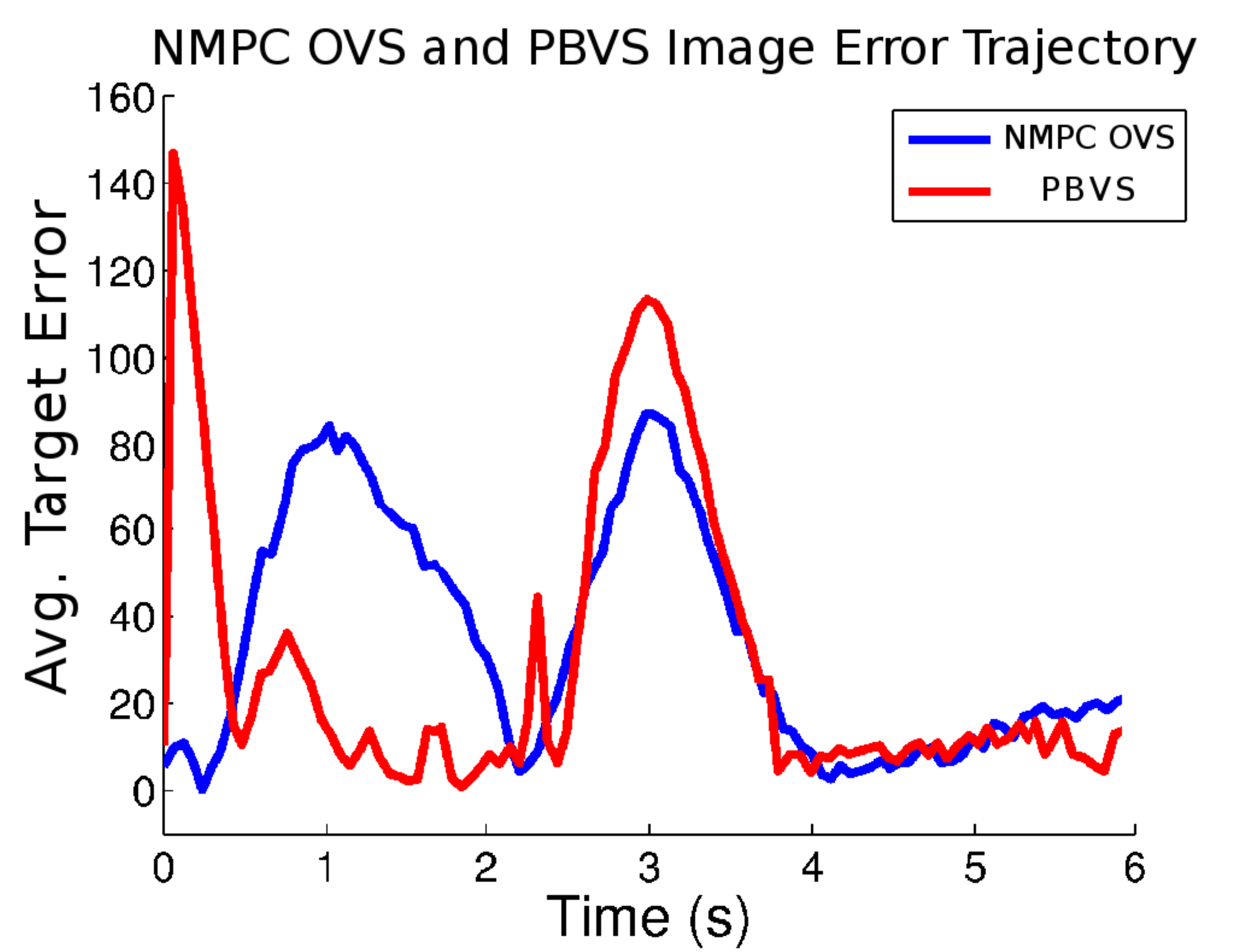}}         
 			{\includegraphics[width=0.49\linewidth]{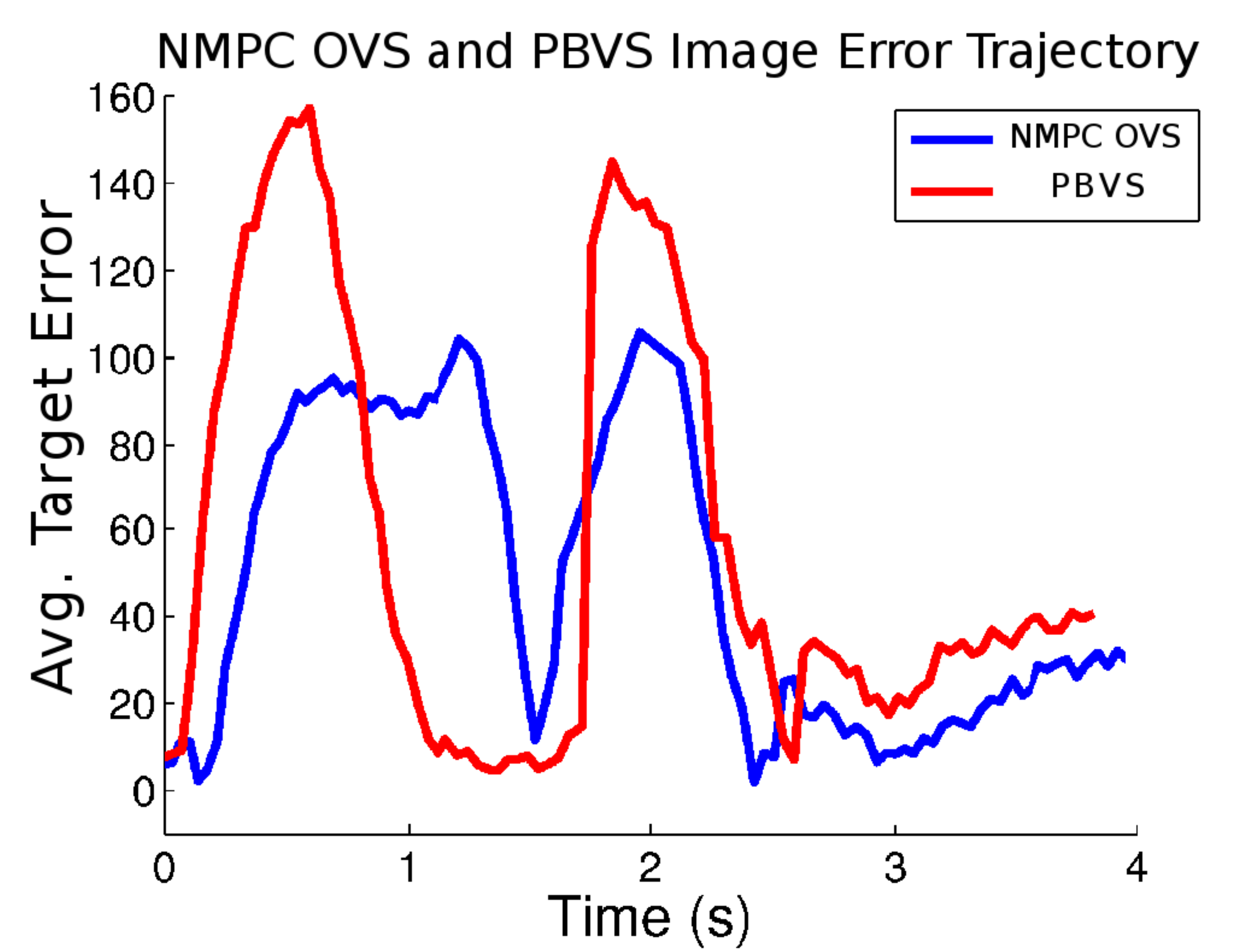}} 
	\caption{\small  Comparison of real PBVS and NMPC OVS Mean Squared Error (MSE) pixel trajectories for different values of $s_{nom}$. Respectively the MSE for $s_{nom} = 2.07$  in the left image, while in the right the MSE for $s_{nom} = 3.1$.}
	\label{fig:real_mpcovs}
\end{figure}

\section{Conclusions}

In this work we proposed a novel OVS approach for multirotor vehicles, particularly suitable for agile maneuvers.
The method splits the VS into two different optimization problems. In the first one an image-based cost function is minimized in order to find the best trajectory for the vehicle. The optimal trajectory is then tracked by means of an NMPC controller that runs in real-time.  
It has been shown in simulated and real-world experiments how the proposed approach achieves better performance in terms of target re-projection error 
when compared to a standard PBVS approach. 
In both scenarios it is capable to keep the target object as close as possible to the center of the camera field of view, even when performing fast maneuvers.

As a future work, we will investigate the possibility to iteratively recompute the optimal trajectory to follow, in order to cope also with the case of a moving target and eventually take into account obstacles in the surrounding environment.

\balance

	\bibliographystyle{IEEEtran}
	\bibliography{IEEEabrv,relwork}

\begin{thebibliography}{10}
\providecommand{\url}[1]{#1}
\csname url@rmstyle\endcsname
\providecommand{\newblock}{\relax}
\providecommand{\bibinfo}[2]{#2}
\providecommand\BIBentrySTDinterwordspacing{\spaceskip=0pt\relax}
\providecommand\BIBentryALTinterwordstretchfactor{4}
\providecommand\BIBentryALTinterwordspacing{\spaceskip=\fontdimen2\font plus
\BIBentryALTinterwordstretchfactor\fontdimen3\font minus
  \fontdimen4\font\relax}
\providecommand\BIBforeignlanguage[2]{{%
\expandafter\ifx\csname l@#1\endcsname\relax
\typeout{** WARNING: IEEEtran.bst: No hyphenation pattern has been}%
\typeout{** loaded for the language `#1'. Using the pattern for}%
\typeout{** the default language instead.}%
\else
\language=\csname l@#1\endcsname
\fi
#2}}

\bibitem{Lee_2012}
D.~Lee, T.~Ryan, and H.~J. Kim, ``Autonomous landing of a {VTOL} {UAV} on a
  moving platform using image-based visual servoing.'' in \emph{Proc. of the
  {IEEE} International Conference on Robotics and Automation ({ICRA})}, 2012.

\bibitem{Falanga2017}
D.~Falanga, M.~Mueggler, E.~Faessler, and D.~Scaramuzza, ``Aggressive quadrotor
  flight through narrow gaps with onboard sensing and computing using active
  vision,'' in \emph{Proc. of the {IEEE} International Conference on Robotics
  and Automation ({ICRA})}, 2017.

\bibitem{Thomas_2014}
J.~Thomas, G.~Loianno, K.~Sreenath, and V.~Kumar, ``{Toward Image Based Visual
  Servoing for Aerial Grasping and Perching},'' in \emph{Proc. of the {IEEE}
  International Conference on Robotics and Automation ({ICRA})}, 2014.

\bibitem{Pestana_2014}
J.~Pestana, J.~L. Sanchez-Lopez, S.~Saripalli, and P.~C. Cervera, ``Computer
  vision based general object following for gps-denied multirotor unmanned
  vehicles.'' in \emph{Proc. of the American Control Conference (ACC)}, 2014.

\bibitem{hamel_2002}
T.~Hamel and R.~Mahony, ``Visual servoing of an under-actuated dynamic
  rigid-body system: an image-based approach,'' \emph{IEEE Transactions on
  Robotics and Automation}, vol.~18, no.~2, pp. 187--198.

\bibitem{Guenard_2007}
N.~Guenard, T.~Hamel, and R.~E. Mahony, ``A practical visual servo control for
  a unmanned aerial vehicle.'' in \emph{Proc. of the {IEEE} International
  Conference on Robotics and Automation ({ICRA})}, 2007.

\bibitem{Altug_2002}
E.~Altug, J.~P. Ostrowski, and R.~Mahony, ``Control of a quadrotor helicopter
  using visual feedback,'' in \emph{Proceedings 2002 IEEE International
  Conference on Robotics and Automation}, 2002.

\bibitem{Mejías_2006}
L.~Mejias, S.~Saripalli, P.~Campoy, and G.~S. Sukhatme, ``Visual servoing of an
  autonomous helicopter in urban areas using feature tracking,'' \emph{Journal
  of Field Robotics}, vol.~23, no. 3-4, pp. 185--199, 2006.

\bibitem{Ozawa_2011}
R.~Ozawa and F.~Chaumette, ``Dynamic visual servoing with image moments for a
  quadrotor using a virtual spring approach,'' in \emph{Proc. of the {IEEE}
  International Conference on Robotics and Automation ({ICRA})}, 2011.

\bibitem{Jabbari_2014}
H.~B. H.~Jabbari, G.~Oriolo, ``Output feedback image-based visual servoing
  control of an underactuated unmanned aerial vehicle,'' in \emph{Proceedings
  of the Institution of Mechanical Engineers, Part I: Journal of Systems and
  Control Engineering}, 2014.

\bibitem{Bourquardez_2009}
O.~Bourquardez, R.~E. Mahony, N.~Guenard, F.~Chaumette, T.~Hamel, and L.~Eck,
  ``Image-based visual servo control of the translation kinematics of a
  quadrotor aerial vehicle,'' \emph{{IEEE} Trans. Robotics}, vol.~25, no.~3,
  pp. 743--749, 2009.

\bibitem{Mebarki_2015}
R.~Mebarki, V.~Lippiello, and B.~Siciliano, ``Autonomous landing of rotary-wing
  aerial vehicles by image-based visual servoing in gps-denied environments,''
  in \emph{Proc. of {IEEE} International Symposium on Safety, Security, and
  Rescue Robotics, {SSRR}}, 2015.

\bibitem{Hafez_2008}
A.~H.~A. Hafez, E.~Cervera, and C.~V. Jawahar, ``Hybrid visual servoing by
  boosting ibvs and pbvs,'' in \emph{Proc. of the 3rd International Conference
  on Information and Communication Technologies: From Theory to Applications},
  2008.

\bibitem{Sheckells_2016}
M.~Sheckells, G.~Garimella, and M.~Kobilarov, ``Optimal visual servoing for
  differentially flat underactuated systems,'' in \emph{Proc. of the {IEEE/RSJ}
  International Conference on Intelligent Robots and Systems ({IROS})}, 2016.

\bibitem{Mahony_2012}
R.~Mahony, V.~Kumar, and P.~Corke, ``Multirotor aerial vehicles: Modeling,
  estimation, and control of quadrotor,'' \emph{IEEE Robotics Automation
  Magazine}, vol.~19, no.~3, pp. 20--32, 2012.

\bibitem{Omari_2016}
S.~Omari, M.~D. Hua, G.~Ducard, and T.~Hamel, ``Nonlinear control of {VTOL}
  {UAVs} incorporating flapping dynamics,'' in \emph{2013 IEEE/RSJ
  International Conference on Intelligent Robots and Systems}, 2013.

\bibitem{BUrri_2012}
M.~Burri, J.~Nikolic, C.~Hürzeler, G.~Caprari, and R.~Siegwart, ``Aerial
  service robots for visual inspection of thermal power plant boiler systems,''
  in \emph{Proc. of the 2nd International Conference on Applied Robotics for
  the Power Industry (CARPI)}, 2012.

\bibitem{Kamel_2016}
M.~Kamel, M.~Burri, and R.~Siegwart, ``Linear vs nonlinear {MPC} for trajectory
  tracking applied to rotary wing micro aerial vehicles,''
  \emph{arXiv:1611.09240}, 2016.

\bibitem{Petersen_2008}
T.~Petersen, ``A comparison of {2D-3D} pose estimation methods,'' \emph{Aalborg
  University, Aalborgb}, 2008.

\bibitem{Furrer_2016}
F.~Furrer, M.~Burri, M.~Achtelik, and R.~Siegwart, \emph{Robot Operating System
  (ROS): The Complete Reference (Volume 1)}, 2016, ch. RotorS---A Modular
  Gazebo MAV Simulator Framework, pp. 595--625.

\bibitem{neunert_2016}
M.~Neunert, C.~de~Crousaz, F.~Furrer, M.~Kamel, F.~Farshidian, R.~Siegwart, and
  J.~Buchli, ``Fast nonlinear model predictive control for unified trajectory
  optimization and tracking,'' in \emph{Proc. of the {IEEE} International
  Conference on Robotics and Automation ({ICRA})}, 2016.

\bibitem{Houska_2011}
B.~Houska, H.~Ferreau, and M.~Diehl, ``{ACADO} {T}oolkit -- {A}n {O}pen
  {S}ource {F}ramework for {A}utomatic {C}ontrol and {D}ynamic
  {O}ptimization,'' \emph{Optimal Control Applications and Methods}, vol.~32,
  pp. 298--312, 2011.

\bibitem{Olson_2011}
E.~Olson, ``{AprilTag}: A robust and flexible visual fiducial system,'' in
  \emph{Proc. of the {IEEE} International Conference on Robotics and Automation
  ({ICRA})}, 2011.

\bibitem{nikolic_2014}
J.~Nikolic, J.~Rehder, M.~Burri, P.~Gohl, S.~Leutenegger, P.~T. Furgale, and
  R.~Siegwart, ``A synchronized visual-inertial sensor system with fpga
  pre-processing for accurate real-time slam,'' in \emph{Proc. of the {IEEE}
  International Conference on Robotics and Automation ({ICRA})}, 2014.

\bibitem{bloesch_2015}
M.~Bloesch, S.~Omari, M.~Hutter, and R.~Siegwart, ``Robust visual inertial
  odometry using a direct {EKF}-based approach,'' in \emph{Proc. of the
  {IEEE/RSJ} International Conference on Intelligent Robots and Systems
  ({IROS})}, 2015.

\bibitem{lynen_2013}
S.~Lynen, M.~Achtelik, S.~Weiss, M.~Chli, and R.~Siegwart, ``A robust and
  modular multi-sensor fusion approach applied to mav navigation,'' in
  \emph{Proc. of the {IEEE/RSJ} International Conference on Intelligent Robots
  and Systems ({IROS})}, 2013.

\end{thebibliography}

\end{document}